\documentclass[runningheads]{llncs}
\usepackage[T1]{fontenc}
\usepackage{graphicx}
\usepackage[misc]{ifsym}
\usepackage{multirow}
\usepackage{wrapfig}
\usepackage{booktabs}
\usepackage{subfigure}
\usepackage{hyperref}[colorlinks,linkcolor=blue]
\usepackage{color}
\renewcommand\UrlFont{\color{blue}\rmfamily}
\usepackage{tikz}
\usetikzlibrary{calc}

\newcommand*\blackcircled[1]{\tikz[baseline=(char.base)]{
        \node[shape=circle,fill={rgb,255:red,0;green,0;blue,0}, text=white, font=\small, inner sep=0.6pt] (char) {#1};}}

\begin{document}
\title{GNNSampler: Bridging the Gap between Sampling Algorithms of GNN and Hardware}
\titlerunning{GNNSampler: Bridging GNN Sampling and Hardware}
% If the paper title is too long for the running head, you can set
% an abbreviated paper title here
%
\author{Xin Liu\inst{1,2} \and
Mingyu Yan (\Letter) \inst{1} \and
Shuhan Song\inst{1,2} \and
Zhengyang Lv\inst{1,2} \and \\
Wenming Li\inst{1,2} \and
Guangyu Sun\inst{3} \and
Xiaochun Ye\inst{1,2} \and
Dongrui Fan\inst{1,2}
}
\authorrunning{X. Liu et al.}
% First names are abbreviated in the running head.
% If there are more than two authors, 'et al.' is used.
%
\institute{
SKLP, Institute of Computing Technology, CAS \and
University of Chinese Academy of Sciences \and
School of Integrated Circuits, Peking University
\email{\{liuxin19g,yanmingyu,songshuhan19s,lvzhengyang19b,\\liwenming,yexiaochun,fandr\}@ict.ac.cn} \\
\email{gsun@pku.edu.cn}
}

\maketitle 

\begin{abstract}
Sampling is a critical operation in Graph Neural Network (GNN) training that helps reduce the cost. Previous literature has explored improving sampling algorithms via mathematical and statistical methods. However, there is a gap between sampling algorithms and hardware. Without consideration of hardware, algorithm designers merely optimize sampling at the algorithm level, missing the great potential of promoting the efficiency of existing sampling algorithms by leveraging hardware features. 
In this paper, we pioneer to propose a unified programming model for mainstream sampling algorithms, termed GNNSampler, covering the critical processes of sampling algorithms in various categories. Second, to leverage the hardware feature, we choose the data locality as a case study, and explore the data locality among nodes and their neighbors in a graph to alleviate irregular memory access in sampling. Third, we implement locality-aware optimizations in GNNSampler for various sampling algorithms to optimize the general sampling process. Finally, we emphatically conduct experiments on large graph datasets to analyze the relevance among training time, accuracy, and hardware-level metrics. Extensive experiments show that our method is universal to mainstream sampling algorithms and helps significantly reduce the training time, especially in large-scale graphs.

\keywords{Graph neural network \and Sampling algorithms \and Acceleration \and Hardware feature \and Data locality}
\end{abstract}

\section{Introduction} 

%Processing graph data has become complex and challenging due to the explosive growth of graph scale in various fields. 
Motivated by conventional deep learning methods, graph neural networks (GNNs) \cite{gnn} are proposed and have shown remarkable performance in graph learning, bringing about significant improvements in tackling graph-based tasks \cite{gcn,app,graphsage}. Whereas, a crucial issue is that real-world graphs are extremely large. Learning large-scale graphs generally requires massive computation and storage resources in practice, leading to high cost in training GNN \cite{sampling_acceleration}. To this end, sampling algorithms are proposed for efficient GNN training, by conditionally selecting nodes to reduce the computation and storage costs in GNN training.
%By conditionally selecting nodes, sampling algorithms reduce the computation and storage costs in GNN training. 

However, abundant irregular memory accesses to neighbors of each node are generally required in all sampling algorithms, introducing significant overhead due to the irregular connection pattern in graph \cite{Yan_Alleviating}. Previous sampling-based models \cite{graphsage,vrgcn,fastgcn,asgcn,clustergcn,graphsaint,PGS-GNN,decoupling} leverage mathematical and statistical approaches for improvement, but they do not alleviate the high cost caused by irregular memory access. Thereby, algorithm designers merely optimize sampling at the algorithm level without considering hardware features. The efficient execution of sampling, and even the efficiency of GNN training, are limited by the gap between algorithm and hardware.

\begin{figure}[t]
\centering
\includegraphics[width=0.42\columnwidth]{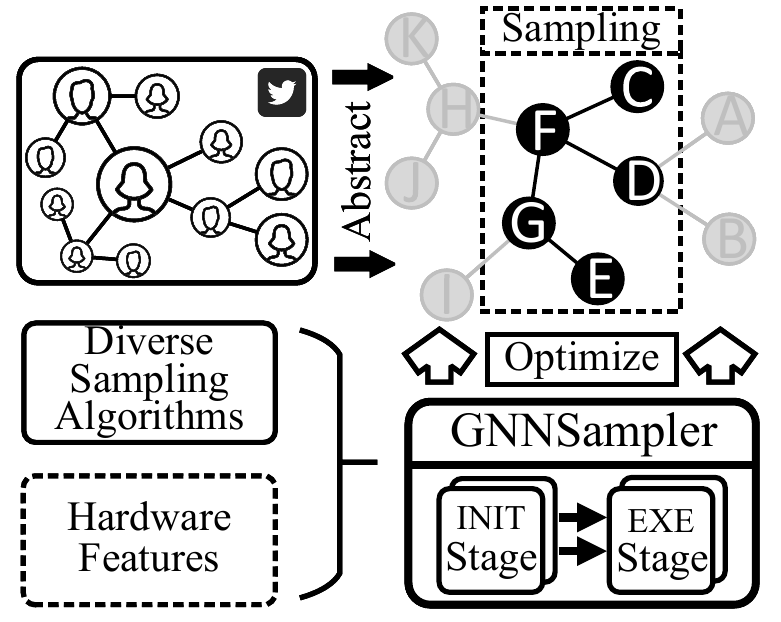}
\caption{Overview of efficient sampling in GNN training based on GNNSampler, where GNNSampler consists of multiple steps in the INIT and EXECUTE stage, making it possible that some substeps of sampling can be optimized in a fine-grained manner by leveraging hardware features.}
\label{fig:gnnsampler_overview}
\end{figure}

To this end, we target to bridge the gap between sampling algorithms of GNN and hardware. In this paper, as illustrated in Figure~\ref{fig:gnnsampler_overview}, we pioneer to build GNNSampler, a unified programming model for sampling, by abstracting diverse sampling algorithms. Our contributions can be summarized as follows:
\begin{itemize}
\item[$\bullet$] We propose a unified programming model, termed GNNSampler, for mainstream sampling algorithms, which covers key procedures in the general sampling process.
\item[$\bullet$] We choose data locality in graph datasets as a case study to leverage hardware features. Moreover, we explore the data locality among nodes and their neighbors in a graph to alleviate irregular memory access in sampling.
\item[$\bullet$] We implement locality-aware optimizations in GNNSampler to improve the general sampling process, helping reduce considerable cost in terms of time. Notably, the optimization is adjustable and is performed once and for all, providing vast space for trading off the training time and accuracy.
\item[$\bullet$] We conduct extensive experiments on large graph datasets, including time-accuracy comparison and memory access quantification, to analyze the relevance among the training time, accuracy, and hardware-level metrics.
\end{itemize} 

\section{Background and Motivation}
In this section, we first introduce the background of GNN and mainstream sampling algorithms. Then, we highlight the gap between the algorithms and hardware, and put forward our motivation.

\subsection{Background of GNN}
GNN \cite{gnn} was first proposed to apply neural networks to graph learning. It learns a state embedding $\textbf{h}_v$ to represent neighborhood information of each node \textit{v} in a graph. Generally, the $\textbf{h}_v$ can be represented in the following form:
\begin{equation}
    \textbf{h}_v = f_{\textbf{w}}(\textbf{x}_v, \textbf{x}_{e[v]}, \textbf{h}_{n[v]}, \textbf{x}_{n[v]}),
\end{equation}
\begin{equation}
    \textbf{o}_{v} = g_{\textbf{w}}(\textbf{h}_{v}, \textbf{x}_{v}),
\end{equation}
where $ \textbf{x}_v$, $\textbf{x}_{e[v]}$, and $\textbf{x}_{n[v]}$ denote the features of $v$, $v$'s edges, and $v$'s neighbors, respectively. $f_{\textbf{w}}(\cdot)$ and $g_{\textbf{w}}(\cdot)$ are the functions defined for local transition and local output. And $\textbf{o}_{v}$ is the output generated by the embedding and feature of node $v$. 
In this way, the hidden information of a graph is extracted by the following approach: 
\begin{equation}
    \textbf{h}^{l+1} = f_{\textbf{w}}(\textbf{h}^{l}, \textbf{x}),
\end{equation}
where $l$ denotes the $l$-th iteration of the embedding computation.
Many variants take the idea from the original GNN and add some particular mechanisms to modify the models for handling various graph-based tasks. 
Herein, we highlight the form of graph convolutional networks (GCNs) \cite{gcn} since most sampling algorithms are applied to GCNs for efficient model training. Generally, GCNs use a layer-wise propagation rule to calculate the embedding in the following form: 
\begin{equation}
    \textbf{H}^{l+1} = \sigma \left( \widetilde{\textbf{D}}^{-1/2} \widetilde{\textbf{A}} \widetilde{\textbf{D}}^{-1/2} \textbf{H}^{l} \textbf{W}^{l} \right),
\end{equation}
where $\textbf{H}^l$, $\textbf{W}^l$ are the hidden representation matrix and trainable weight matrix in the $l$-th layer of the model. And $\sigma(\cdot)$ is the nonlinear activation function, such as ReLU and Softmax. GCNs represent neighborhood information with a renormalized adjacency matrix and extract the hidden information in such an iterative manner.

\begin{figure}[t]
\centering
\includegraphics[width=0.7\columnwidth]{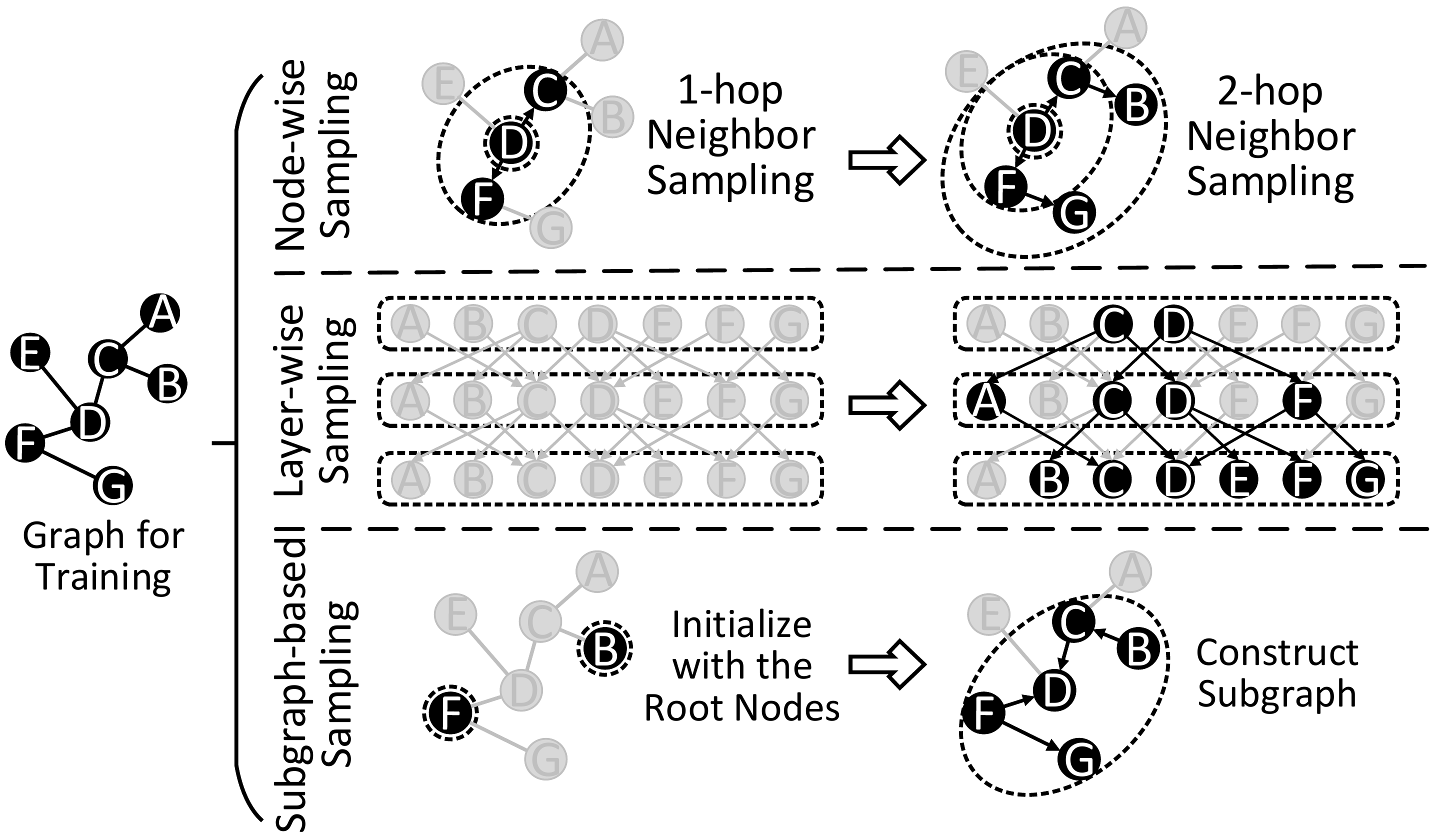}
\caption{Illustration on typical processes of various sampling algorithms.}
\label{3Sampling}
\end{figure}

\subsection{Sampling Algorithms in Training} \label{sec:related work}
Training GNNs, especially GCNs, generally requires full graph Laplacian and all intermediate embeddings, which brings about extensive storage cost and makes it hard to scale the training on large-scale graphs. Moreover, the conventional training approach uses a full-batch scheme to update the model, leading to a slow convergence rate.%, which makes the training inefficient.

To overcome these drawbacks, sampling algorithms are proposed to modify the conventional training through a mini-batch scheme and conditionally select partial neighbors, reducing the cost in terms of storage and computation. Specifically, \textbf{mainstream sampling algorithms} can be divided into multiple categories according to the granularity of the sampling operation in one sampling batch \cite{sampling_survey}. As illustrated in Figure~\ref{3Sampling}, we respectively show typical sampling processes of multiple categories, that is, \textbf{node-wise}, \textbf{layer-wise}, and \textbf{subgraph-based} sampling algorithms.

As typical of \textbf{node-wise} sampling algorithms, GraphSAGE \cite{graphsage} randomly samples the neighbors of each node in multiple hops recursively; VR-GCN \cite{vrgcn} improves the strategy of random neighbor sampling by restricting sampling size to an arbitrarily small value, which guarantees a fast training convergence.
\textbf{Layer-wise} sampling algorithms, e.g., FastGCN \cite{fastgcn} and AS-GCN \cite{asgcn}, generally conduct sampling on a multi-layer model in a top-down manner. FastGCN presets the number of nodes to be sampled per layer without paying attention to a single node's neighbors. AS-GCN executes the sampling process conditionally based on the parent nodes sampled in the upper layer, where the layer sampling is probability-based and dependent among layers. 
As for \textbf{subgraph-based} sampling algorithms, multiple subgraphs, which are generated by partitioning the entire graph or inducing nodes (edges), are sampled for each mini-batch for training. Cluster-GCN \cite{clustergcn} partitions the original graph with a clustering algorithm and then randomly selects multiple clusters to construct subgraphs. GraphSAINT \cite{graphsaint} induces subgraphs from probabilistic sampled nodes (edges) by leveraging multiple samplers. %In summary, all the sampling algorithms select partial nodes based on specific rules and add them into one batch for training. 

\begin{figure}[t]
\centering
\includegraphics[width=0.6\columnwidth]{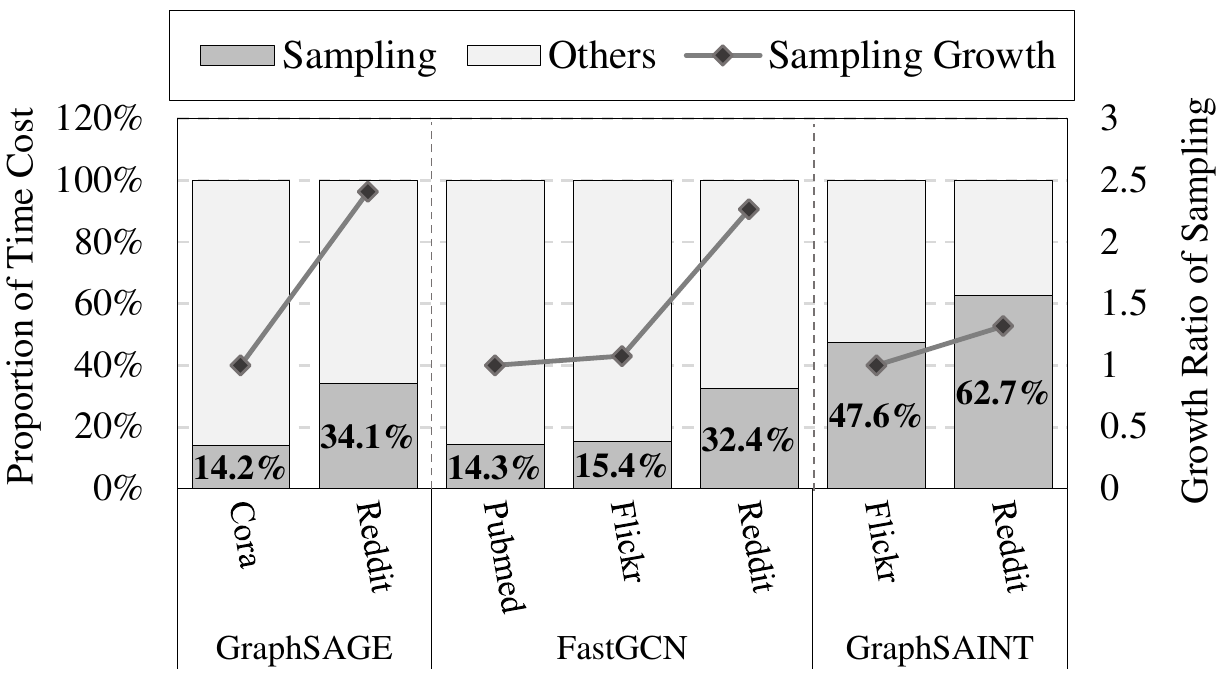}
\caption{The proportion of the execution time in different parts of training across different models and datasets.}
\label{fig:proportion}
\end{figure}

Unfortunately, the cost of sampling is gradually becoming non-negligible in the training of some sampling-based models, especially on large datasets. As proof, we conduct experiments on datasets with a growing graph scale (amount of nodes \& edges) using GraphSAGE \cite{graphsage}, FastGCN \cite{fastgcn} and GraphSAINT (node sampler) \cite{graphsaint}, and quantify the proportion of sampling time to the training time, \textbf{based on the official setting of all parameters in their works}. The sampling part includes selecting nodes, constructing an adjacency matrix, and some subsequent processes. The other part denotes the rest of processes in training, e.g., feature aggregation and update. For each sampling algorithms, we consider the proportion of sampling time in the smallest dataset (e.g., Cora \cite{dataset}) as the baseline and plot the growth trend of sampling time on each dataset. Detailed information about datasets is shown in Table \ref{tab:dataset}. Distinctly, for all these sampling-based models, as the number of nodes and edges in a graph dataset grows, the cost of sampling becomes increasingly large and intolerable. As illustrated in Figure~\ref{fig:proportion}, the proportion of sampling time becomes larger as the graph scale of a dataset increases, even up to 62.7\%.

\subsection{The Gap between Algorithm and Hardware}  \label{sec:motivation}
As a critical process in training, sampling is becoming non-trivial, and its cost mainly derives from the gap between sampling algorithms and hardware. Algorithm designers do not consider hardware features and merely improve sampling at the algorithm level. On the other hand, hardware designers have not improved sampling algorithms since they do not know specific implementations of sampling algorithms. Therefore, mining of the improvement space for sampling algorithms is restricted by the gap.
Thereby, recent literature proposes to leverage the hardware feature for improving the efficiency regarding training and inference of GNNs.
%Moreover, leveraging hardware features of modern architectures or customizing hardware features of architecture designs help improve GNN training and inference efficiently. 
For example, NeuGraph \cite{neugraph} makes the best of the hardware features of platforms (CPU and GPU), achieving excellent improvements in GNN training. 
For another example, HyGCN \cite{hygcn} tailors its hardware features to the execution semantic of GCNs based on the execution characterization of GCNs~\cite{Yan_Characterizing}, greatly improving the performance of GCN inference. 

We argue that sampling is also a process of algorithm and hardware coordination. We observe that existing sampling algorithms vary in their mechanisms. And an efficient improvement should be universal to most sampling algorithms of different mechanisms, urging the demand to put forward a general sampling programming model. To support our argument, we propose GNNSampler, a unified programming model for mainstream sampling algorithms. Moreover, we choose data locality as a case study and implement locality-aware optimizations to improve the general sampling process. By improving the process of sampling, we eventually reduce the time consumption of GNN training.

\section{Unified Programming Model}
In this section, we abstract sampling algorithms in different categories and propose the unified programming model. 
%To make the programming model compatible with different sampling algorithms, we extract the key procedures of sampling algorithms in each category based on their implementations. 

\begin{figure}[t]
\centering
\includegraphics[width=0.99\textwidth]{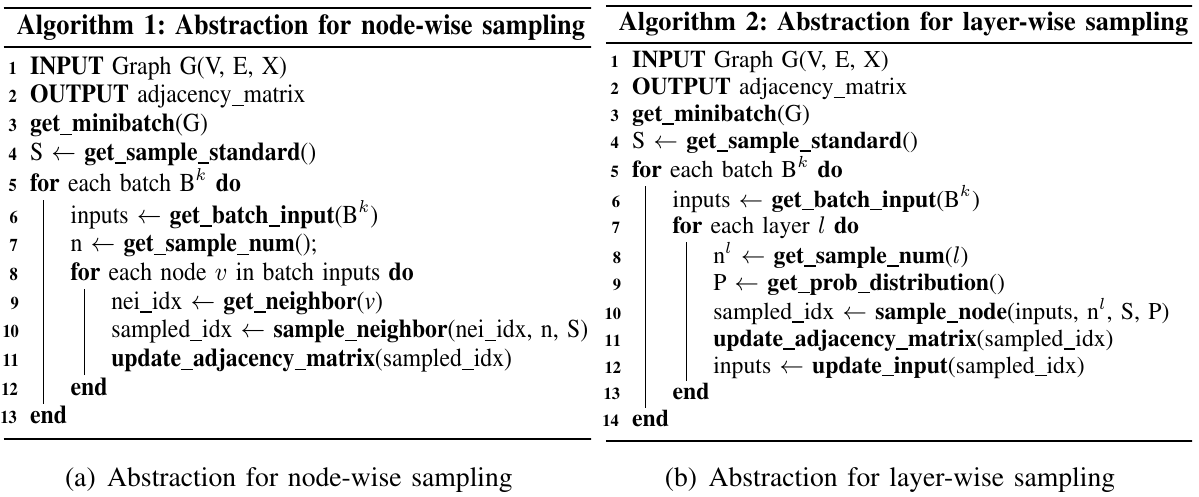}
\caption{(a) Abstraction for node-wise sampling algorithms. (b) Abstraction for layer-wise sampling algorithms. To be as unified as possible, we use the same name for steps (functions) in the abstractions, despite some slight distinctions of required parameters.}
\label{fig:nodeANDlayer}
\end{figure}

\subsection{Abstractions for Sampling Algorithms} 

$\bullet$ \textit{Abstraction for Node-wise Sampling Algorithms.}

In Figure \ref{fig:nodeANDlayer} (a), lines 3-4 of \textbf{Algorithm 1} correspond to the initialization phase of sampling. In this phase, nodes needed for training are first divided into multiple mini-batches. Lines 5-13 correspond to the execution phase of sampling. First, nodes in a mini-batch are first obtained in the form of an indices list. Next up, the sampling size for each node is obtained before sampling. Then, sampling is executed on each node in a batch to get indices of neighbors by leveraging some pre-calculated parameters. Finally, the adjacency matrix is constructed based on the sampled nodes and updated periodically per batch. 

\noindent $\bullet$ \textit{Abstraction for Layer-wise Sampling Algorithms.}

In Figure \ref{fig:nodeANDlayer} (b), layer-wise samplings do not need to focus on a single node since they sample a fixed number of nodes together in each layer based on the pre-calculated probability distribution. 
In \textbf{Algorithm 2}, line 9 corresponds to the calculation of probability distribution. Line 10 represents that n$^l$ nodes are sampled together in the \textit{l}-th layer by leveraging some pre-calculated parameters.
Generally, the indices of sampled nodes are used to update the adjacency matrix. For some layer-wise sampling algorithms (e.g., AS-GCN \cite{asgcn}) in which nodes are sampled according to parent nodes (already sampled nodes) in the upper layer, the indices are further used to update the input of sampling for the next layer.

\noindent $\bullet$ \textit{Abstraction for Subgraph-based Sampling Algorithms.}

\begin{figure}[t]
\centering
\includegraphics[width=0.99\columnwidth]{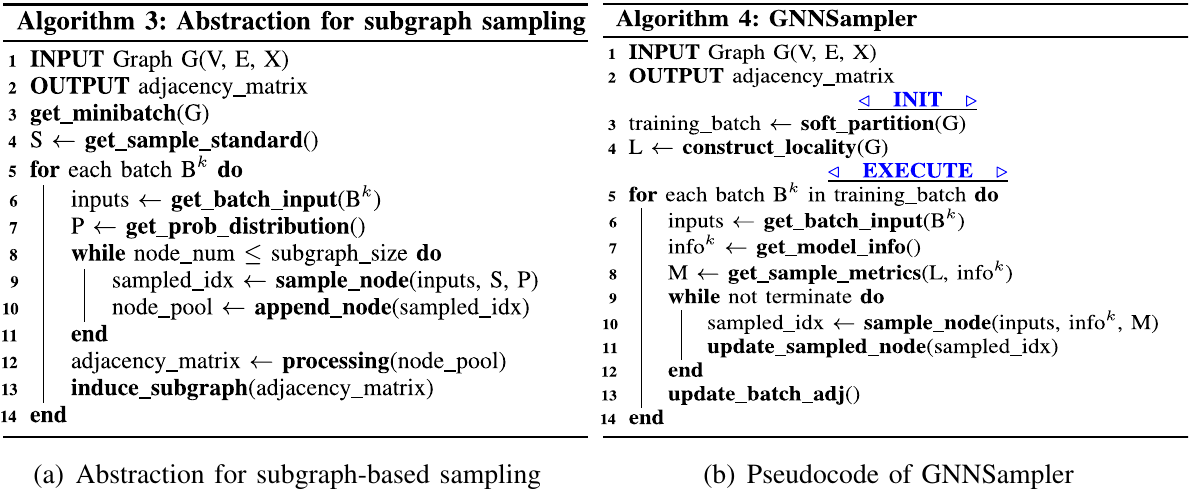}
\caption{(a) Abstraction for subgraph-based sampling algorithms. (b) Pseudocode of GNNSampler. Please note that, in subplot (b), we add a particular step, termed ``construct\_locality'', to the INIT stage to exploit the data locality among nodes and their neighbors in a graph. In the EXECUTE stage, sampling can be executed with less irregular memory access since the sampling weight is computed according to the data locality among nodes.}
\label{fig:subANDgnnsampler}
\end{figure}

In Figure \ref{fig:subANDgnnsampler} (a), mini-batches obtained in line 3 of \textbf{Algorithm 3} correspond to the training nodes partitioned artificially or with a clustering algorithm. Lines 8-11 correspond to the main loop of sampling, and it does not stop until the number of sampled nodes equals the number needed to form a subgraph. The sampled nodes are temporarily stored in a ``node\_pool''. The function named ``processing'' in line 12 denotes a further process after sampling, e.g., sorting nodes' indices and removing the duplicate nodes. Finally, a subgraph is induced from the adjacency matrix generated by sampled nodes. Multiple subgraphs are generated by repeating the above process.

\subsection{GNNSampler and Workflow}

\begin{figure}[t]
\centering
\includegraphics[width=0.99\textwidth]{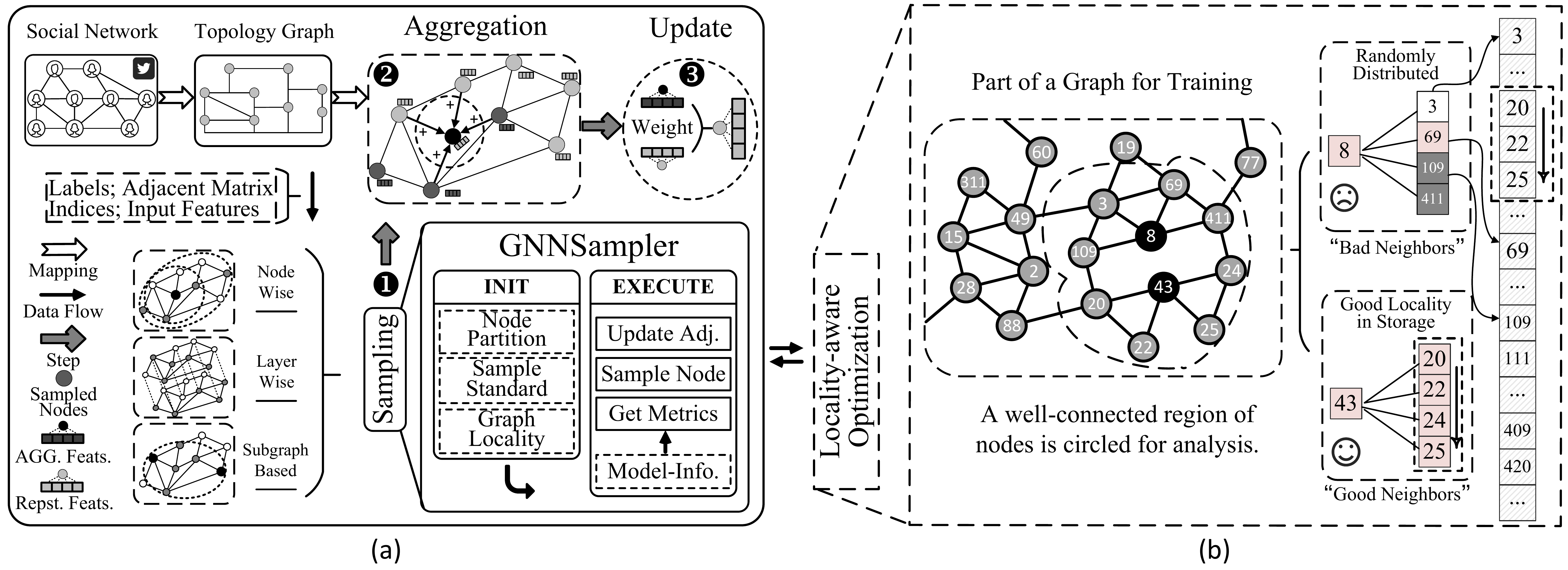} 
\caption{(a) Workflow of learning large-scale graphs with GNN embedded with GNNSampler (some general processes, e.g., loss compute and model update, are not specially shown). (b) An exemplar for sampling nodes leveraging hardware feature.}
\label{framework}
\end{figure}

\textbf{Design:} Based on the above abstractions, we propose the unified programming model, i.e. GNNSampler, in Figure~\ref{fig:subANDgnnsampler} (b). We divide the sampling process into two stages, namely INIT and EXECUTE. The target of the INIT stage is to obtain necessary data, e.g., batched nodes, in advance for the EXECUTE stage. In the EXECUTE stage, line 7 denotes that information of model structure is obtained to help configure the sampling. In line 8, the obtained metrics denote critical factors for sampling, e.g., sampling size and probability. We also add the influence of data locality in the calculation of the sampling probability (i.e., L that computed by function construct\_locality). Lines 9-12 denote an iterative sampling process requiring significant computation and storage resources. Finally, the batched adjacency matrix is updated after the batched sampling. And subsequent steps, e.g., subgraphs induction and model training, can directly use the generated adjacency matrix.

\iffalse
\begin{figure*}[t]
\begin{minipage}[t]{0.31\linewidth}
\centering
\includegraphics[width=0.95\columnwidth]{FIG/gnnsampler1.pdf}
\caption{Pseudocode \\ of GNNSampler.}
\label{fig:gnnsampler_psd}
\end{minipage}%
\begin{minipage}[t]{0.69\linewidth}
\centering
\includegraphics[width=0.95\textwidth]{FIG/framework-v5.pdf} 
\caption{(a) Workflow of learning large-scale graphs with GNN embedded with GNNSampler(some general processes, e.g., loss compute and model update, are not specially shown). (b) An exemplar for sampling nodes leveraging hardware feature.}
\label{framework}
\end{minipage}
\end{figure*}
\fi

\noindent \textbf{Workflow:} 
To embed GNNSamlper to GNN training, we first introduce the steps of pre-processing, sampling, aggregation, and update in GNN training, where sampling is a tight connection between other steps. Figure 6 (a) illustrates the workflow of learning large-scale graphs with GNN, where GNNSampler is embedded for optimizing sampling. 
To begin with, the graph data, such as social network, is transformed into a topology graph and is further converted into elements directly used during sampling, in the pre-processing step. \blackcircled{1} In the sampling step, GNNSampler decomposes sampling into INIT and EXECUTE stages. The processed elements are fed into the INIT stage to compute critical metrics for sampling. In the EXECUTE stage, sampling is performed according to the critical metrics and acquires an adjacency matrix. \blackcircled{2} In the aggregation step, the aggregation feature (abbreviated as AGG. Feats.) of one node is aggregated from features of the sampled nodes, after which a concatenation operation is applied to the AGG. Feats. and the representation feature (abbreviated as Repst. Feats.) of the node in the upper layer. \blackcircled{3} In the update step, the Repst. Feats. of one node is updated by transforming the weighted concatenate feature with a nonlinear function \cite{graphsage}. The most critical one, i.e., the sampling process with GNNSampler embedded, is designed to be universal for all categories of sampling algorithms. Based on this universal design, we can propose a generic and highly compatible optimization to benefit all categories of sampling algorithms.

\section{Case Study: Leveraging Locality}
In this section, to leverage the hardware feature, we choose the data locality as a case study and implement locality-aware optimizations in GNNSampler to improve sampling. Please note that we refer to data locality as locality in brief in the rest of the paper.

%In this section, to leverage the hardware feature, we choose the data locality as a case study and implement locality-aware optimizations in GNNSampler to improve the sampling process. 

\begin{figure}[t]
\centering
\includegraphics[width=0.7\textwidth]{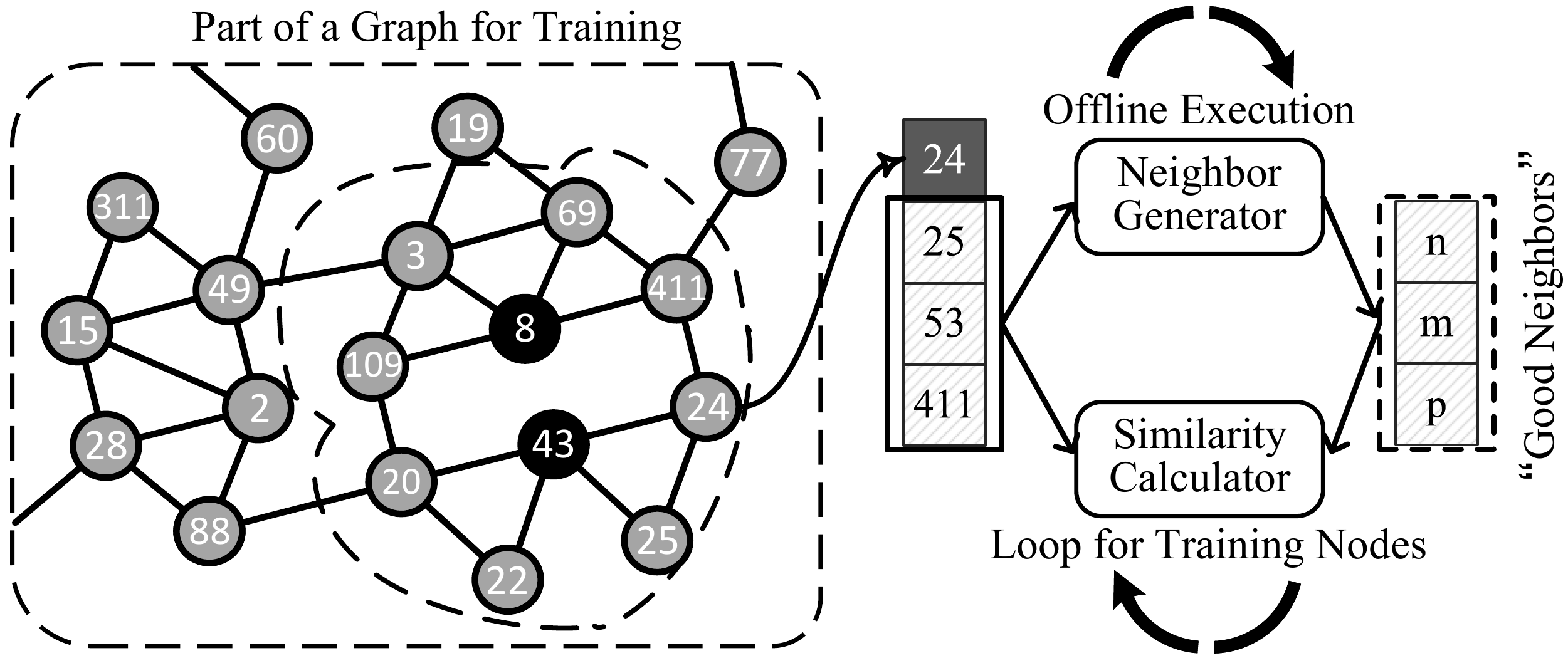} 
\caption{Illustration of two critical steps of the locality-aware optimization: generating good neighbors and calculating the similarity between real and generated good neighbors. Please note that, this process is offline performed for each training node.}
\label{fig:locality-demo}
\end{figure}

\begin{figure}[t] 
\centering
\includegraphics[width=0.8\columnwidth]{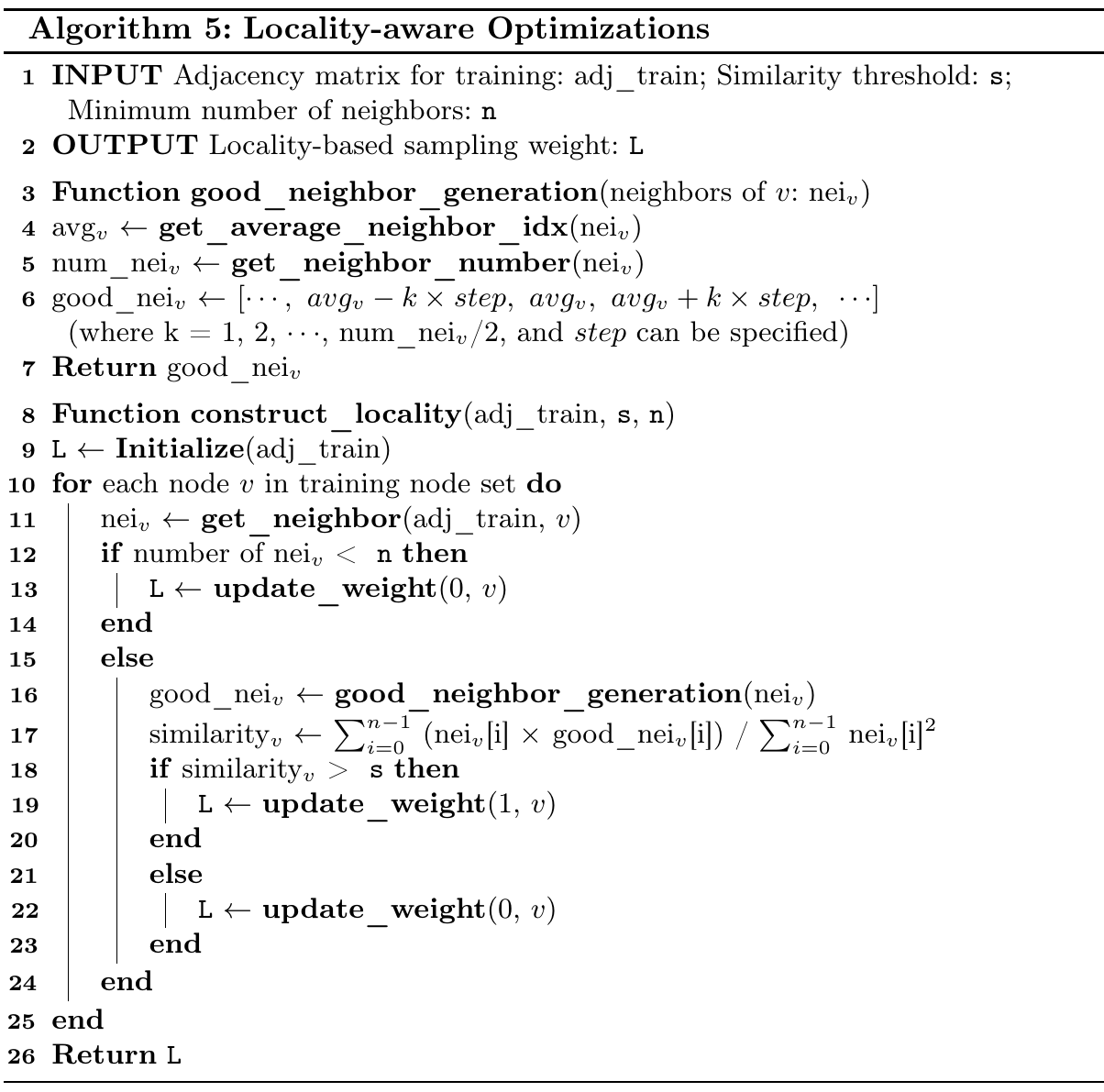}
\caption{Pseudocode of locality-aware optimizations.}
\label{fig:locality}
\end{figure} 

\subsection{Exploring Locality in Graph} \label{sec:Locality}
Locality, i.e., principle of locality \cite{locality}, is a particular tendency that processors have more opportunity to access data in the same set of memory locations repeatedly during a shorter interval of time.  Specifically, locality can be embodied in two basic types: temporal locality for denoting the repeated access to the same memory location within a short time, and spatial locality for denoting the access to particular data within close memory locations.

Inspired by the success of locality exploitation in graph processing~\cite{explore_locality}, we exploit locality to alleviate the irregular memory access in the sampling which helps reduce the execution time. Figure \ref{framework} (b) gives an exemplar for analyzing neighbors of two nodes (i.e., ``No.8'' and ``No.43'') in a well-connected region. Neighbors of node ``No.8'' are randomly distributed, whilst neighbors of node ``No.43'' are almost contiguous. Notably, neighbors of ``No.43'' have a greater probability of being accessed within a short time according to the feature of locality since they are stored in adjacent location in memory. We would sample node ``No.43'' rather than ``No.8''.

Moreover, a new graph is constructed before sampling ends based on the sampled nodes in some cases. In this process, all nodes for training are required to search whether their neighbors are stored in the pool of the sampled nodes. For a node like ``No.43'', it is immediate to search the adjacent locations in the pool to verify their connections and add connections in storage (e.g., CSR format). When the graph is large, locality-aware optimizations can reduce considerable time cost in the above process since irregular memory accesses in searching neighbors are avoided as possible.

\subsection{Implementations of Locality-aware Optimization} \label{sec:4.2}
\textbf{Design flow:} 
The distribution of ``No.43'' and its neighbors is an exactly ideal case in a graph. In most situations, nodes are randomly distributed in general. Therefore, to estimate which node is suitable for sampling, we design two modules, i.e., a generator to yield (virtual) good neighbors and a calculator to compute the similarity between real neighbors and the generated good neighbors. As illustrated in Figure \ref{fig:locality-demo}, for one node $v$ used in training, based on our prescript of good neighbors, i.e., neighbors are contiguous and adjacently located in memory, we use a generator to yield good neighbors for the given $v$. However, there may exist a gap between real neighbors and the generated good neighbors. We then utilize a calculator to quantify the gap. Specifically, we calculate the similarity between two neighbor sequences via a dot product ratio scheme:
\begin{equation} 
    Similarity_v = \frac{\sum_{i=0}^{n-1} n_v[i] \cdot gn_v[i]}{\sum_{i=0}^{n-1} n_v^2[i]}
\end{equation} 
where $n_v$ and $gn_v$ denote sequences of real and the generated good neighbors of $v$. Through evaluation, we discover that the calculated similarity is generally larger as the resemblance of sequences' distributions increases. By setting a suitable threshold for the similarity, nodes whose real neighbors meet our standard (i.e., exceed the threshold) are chosen for sampling. The sampling process is uniformly performed (i.e., the sampling probability is uniform) on these nodes.

\noindent \textbf{Implementation:} 
We implement locality-aware optimizations for all categories of sampling algorithms. For node-wise and subgraph-based sampling algorithms, we use the function ``construct\_locality'' given in Figure \ref{fig:locality} to construct locality among nodes and generate sampling weight. The input are the adjacency matrix, the minimum number of neighbors per node (abbreviated as \texttt{n}), and the similarity threshold (abbreviated as \texttt{s}). 
\texttt{n} is used to filter nodes with sparse connections, and \texttt{s} is used to measure the quality of locality among neighbors of each node by comparing the similarity between real neighbors and the generated good neighbors. The function ``good\_neighbor\_generation'' is used to generate neighbor. The output of the function ``construct\_locality'' is the locality-based sampling weight \texttt{L} filled with ``0'' or ``1'' value which is used to represent whether a node $v$ is suitable to be sampled. The subsequent sampling is performed based on \texttt{L}. 
By this means, nodes whose neighbors are stored closely in memory (i.e., with good locality) are more likely to be sampled, helping alleviate irregular data access.
For layer-wise sampling algorithms, number of nodes to be sampled in each layer is relatively small and fixed for all datasets, which tends to form sparse connections between layers, especially in large datasets. We thereby explore leveraging the number of neighbors and previously sampled nodes to construct locality in two continuous layers. We first initialize a weight vector \texttt{L} for training nodes and set $l_v$ as the weight in the corresponding position in \texttt{L}, where $l_v$ is directly proportional to the number of $v$'s neighbors. Moreover, nodes sampled in the upper layer are partly added to the candidate set to be sampled in the current layer to increase the sampling probability of the frequently accessed nodes.

\noindent \textbf{Once-for-all:} The proposed optimization is flexible to be embedded in the pre-processing step of mainstream sampling-based models. Moreover, two parameters, i.e., \texttt{n} and \texttt{s}, can be adaptively adjusted to achieve the desired trade-off. Notably, the computation of \texttt{L} merely requires the connections among training nodes and their neighbors (i.e., an adjacency matrix for training: adj\_train) and \textbf{can be performed offline}. The pre-computed \texttt{L} can be reused in each batch of sampling, making the computation of \texttt{L} a \textbf{once-for-all process for each dataset}. 
Please refer to our code \footnote[1]{\UrlFont{https://github.com/TeMp-gimlab/GNNSampler}} for more details.
%to use the optimization for accelerating the sampling process.

\section{Experiment}
To analyze the advance of our method, we conduct experiments on all categories of sampling algorithms to compare \textbf{vanilla} training methods (i.e., original models) and \textbf{our} improved approaches with locality-aware optimizations. 
%In this section, we conduct extensive experiments on all categories of sampling algorithms to compare \textbf{vanilla} training methods (i.e., original models) and \textbf{our} improved approaches with locality-aware optimizations. 
%Moreover, we analyze the experimental result to prove the effectiveness and efficiency of our method.

\subsection{Experimental Setup}
Since the GNNSamper is general and compatible with mainstream sampling algorithms, we choose sampling algorithms in all categories as representatives, including GraphSAGE \cite{graphsage} (node-wise sampling), FastGCN \cite{fastgcn} (layer-wise sampling), and GraphSAINT \cite{graphsaint} (subgraph-based sampling). For all sampling-based models, 
\textbf{we use their official configurations in both sampling and training,} especially batch size, sampling size, and learning rate, to guarantee similar performance compared to their reported values. The basic GCN used in all cases is a two-layer model. For GraphSAINT, we choose the serial node sampler implemented via Python to randomly sample nodes for inducing subgraphs.
These sampling-based models are regarded as comparison baselines, and we apply locality-aware optimizations to these models for improvement. By referring to the practice of previous works, we mainly focus on five benchmark datasets distinguishing in graph size and connection density as shown in Table~\ref{tab:dataset}.
All experiments are conducted on a Linux server equipped with dual 14-core Intel Xeon E5-2683 v3 CPUs and an NVIDIA Tesla V100 GPU (16 GB memory).

\begin{figure}[t]
\centering
\subfigure[Comparison of training time]{
\begin{minipage}[t]{0.5\linewidth}
\centering
\includegraphics[width=0.98\columnwidth]{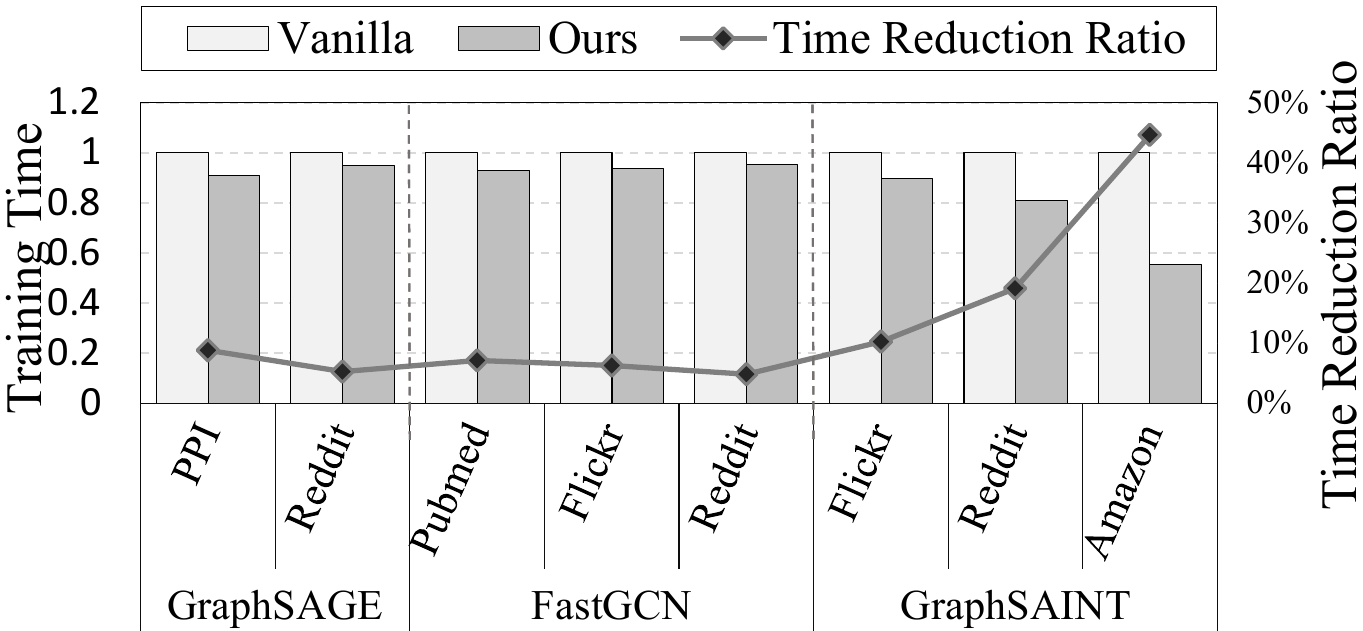}
%\caption{fig1}
\end{minipage}%
}%
\subfigure[Comparison of validation accuracy]{
\begin{minipage}[t]{0.5\linewidth}
\centering
\includegraphics[width=1\columnwidth]{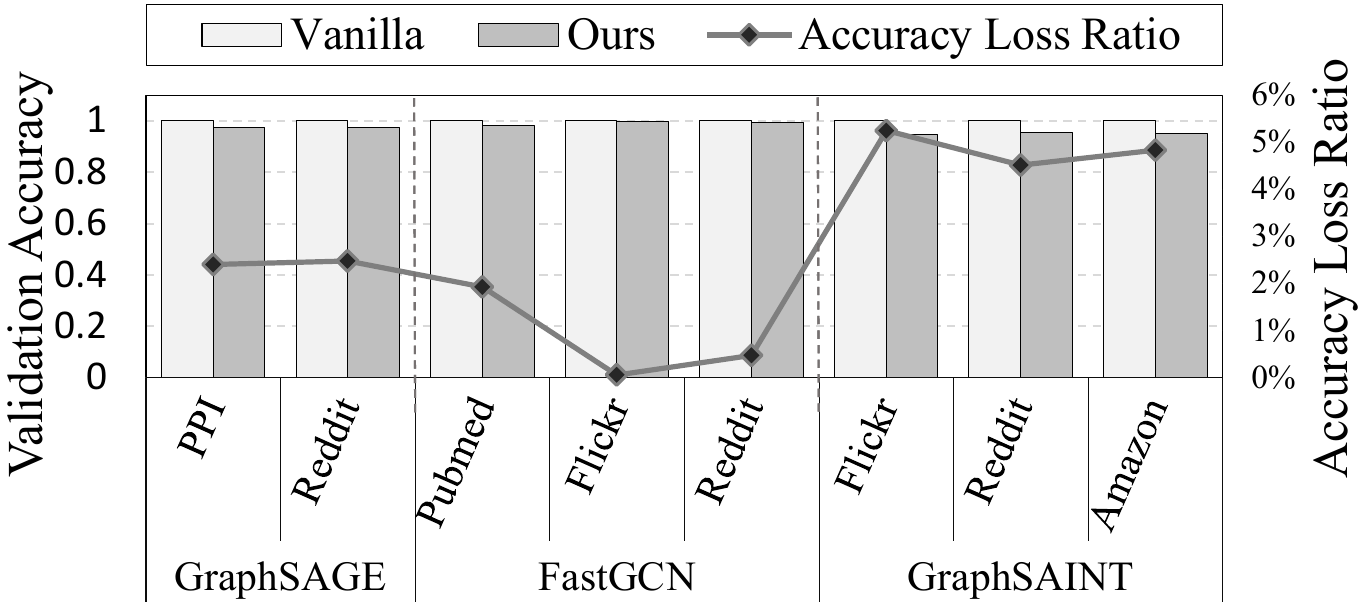}
%\caption{fig2}
\end{minipage}%
}%
\centering
\caption{Comparisons between vanilla and our optimized approaches (normalized to vanilla) among three models using various datasets.}
\label{fig:comparison}
\end{figure}

\tabcolsep 8pt
\begin{table}[t]
\centering
\caption{Comprehensive analysis on the datasets.}
\begin{tabular}{ccccccc}
\bottomrule
\textbf{Dataset} & \textbf{\#Node}  & \textbf{\#Edge} & \textbf{ANN} & \textbf{MNN} & \textbf{NRR} \\ \hline
Pubmed \cite{dataset}     & 19717  & 44338      & 4   & 12  & 77.76\% \\ 
PPI    \cite{dataset_PPI}  & 14755  & 225270     & 15  & 39  & 93.54\% \\ 
Flickr \cite{graphsaint}        & 89250  & 899756     & 5   & 9   & 80.16\% \\ 
Reddit \cite{graphsage} & 232965 & 11606919   & 50  & 113 & 97.99\% \\ 
Amazon \cite{graphsaint}       & 1598960 & 132169734  & 85  & 101 & 98.82\% \\ 
\bottomrule
\end{tabular}
\label{tab:dataset} 
\end{table}

\subsection{Experimental Result and Analysis}

\textbf{Preliminary:} We first analyze the datasets in multiple aspects. As given in Table \ref{tab:dataset}, we make statistics on the (round-off) average number of neighbors per node (\textbf{ANN}) and the maximum number of neighbors of 90\% nodes (\textbf{MNN}) in datasets to reflect density of connections among nodes. We count neighbor reusing rate (\textbf{NRR}) by calculating the number of reused neighbors as a proportion of the total number of neighbors of all nodes. The collected statistics will help establish a relationship among attributes (e.g., size, density of connection) of graph datasets and experimental results.

\begin{table}[ht]
\centering
\caption{Training time and accuracy comparisons on different sampling algorithms between vanilla and optimized approaches. Please note that, parameters \texttt{n} and \texttt{s} denote the minimum number of neighbors per node and the similarity threshold, respectively. Related content has been detailedly discussed in Section \ref{sec:4.2}.}\label{tab:comparison}
\resizebox{\linewidth}{!}{
\begin{tabular}{ccccccc}
\bottomrule
\textbf{Model} & \textbf{Dataset} & \begin{tabular}[c]{@{}c@{}}\textbf{Sampling}\\ \textbf{Size}\end{tabular} & 
\begin{tabular}[c]{@{}c@{}}\textbf{Training Time}\\ \textbf{(Vanilla / Ours)}\end{tabular} &
\begin{tabular}[c]{@{}c@{}}\textbf{Time} \\ \textbf{Reduction} \end{tabular}  &
\begin{tabular}[c]{@{}c@{}} \textbf{Param.} \\ \textbf{(\texttt{n} / \texttt{s})}\end{tabular} &
\begin{tabular}[c]{@{}c@{}}\textbf{Accuracy} \\ \textbf{Loss} \end{tabular}
\\ \hline
\multirow{2}{*}{GraphSAGE} & PPI & 25 \& 10 & 34.97s / \textbf{31.89s} & \textbf{8.81\%} &  4/0.95 & 2.41\%\\ 
 & Reddit & 25 \& 10 & 299.21s / \textbf{283.40s} & \textbf{5.28\%} & 4/0.87 & 2.48\% \\ \hline
\multirow{3}{*}{FastGCN} & Pubmed & 100 & 35.84s / \textbf{33.30s} & \textbf{7.09\%} & - & 1.93\% \\ 
 & Flickr & 100 & 110.29s / \textbf{103.34s} & \textbf{6.30\%} & - & 0.06\%\\ %\cline{2-7ling} 
 & Reddit & 100 & 671.60s / \textbf{639.16s} & \textbf{4.83\%} & - & 0.48\% \\ \hline
\multirow{3}{*}{GraphSAINT} 
 & Flickr & 8000 & 16.04s / \textbf{14.39s} & \textbf{10.29\%} & 3/0.865 & 5.25\% \\
 & Reddit & 4000 & 229.47s / \textbf{185.67s} & \textbf{19.09\%} & 3/0.77 & 4.51\% \\  
 & Amazon & 4500 & 2263.92s / \textbf{1253.33s} & \textbf{44.64\%} & 4/0.775 & 4.83\% \\
\bottomrule
\end{tabular}
}
\end{table}

\noindent \textbf{Result:} As illustrated in Figure \ref{fig:comparison} (a), we compare the converged training time and validation accuracy on diverse sampling-based models and datasets. The overall average time reduction is 13.29\% with a 2.74\% average accuracy loss. Specifically, the average time reduction on GraphSAGE and FastGCN is 7.06\% and 7.14\%.
Notability, the optimization achieves an average 24.67\% time reduction in GraphSAINT, while the peak of time reduction is 44.62\% on Amazon dataset. We also observe a trivial decline in accuracy in Figure \ref{fig:comparison} (b), which varies by model. Considering characteristics of locality, nodes whose neighbors are adjacently distributed have a higher probability of being sampled, which yields a non-uniform sampling distribution. Therefore, biased sampling can result in a sacrifice in accuracy despite the considerable time reduction.
Detailed performance and parameters (\texttt{n} and \texttt{s}) are given in Table \ref{tab:comparison}.

\noindent \textbf{Analysis:} Based on the result, we analyze the relevance among the training time, accuracy, and hardware-level metrics, and summarize our findings as follows:

\noindent $\bullet$ \textbf{Time reduction introduced by the optimization varies by dataset because of the distinct sampling size and graph scale.} Distinctly, the percentage of time reduction in GraphSAINT is larger than in GraphSAGE and FastGCN, since the sampling size used in GraphSAINT is quite large. Moreover, for GraphSAINT, the percentage of time reduction on Amazon dataset is larger than on Flickr and Reddit since Amazon has an enormous graph scale (amount of nodes \& edges). We also argue that higher \textbf{NRR} is another reason for achieving significant time reduction on Amazon. As shown in Table \ref{tab:dataset}, \textbf{NRR} is a metric to reflect neighbor reusing rate.
Generally, a dataset with higher \textbf{NRR} includes node regions in which multiple nodes have many common neighbors. If neighbors of one node are frequently accessed, it is likely that for other nodes in such regions, their neighbors are also frequently accessed since they share many common neighbors. Thus, locality among nodes is more easier to explore in this case. Consequently, models using large sampling sizes and large-scale graphs are more likely to benefit from locality-aware optimizations.

\begin{figure}[t]
\centering
\subfigure[]{
\begin{minipage}[t]{0.74\linewidth}
\centering
\includegraphics[width=0.98\columnwidth]{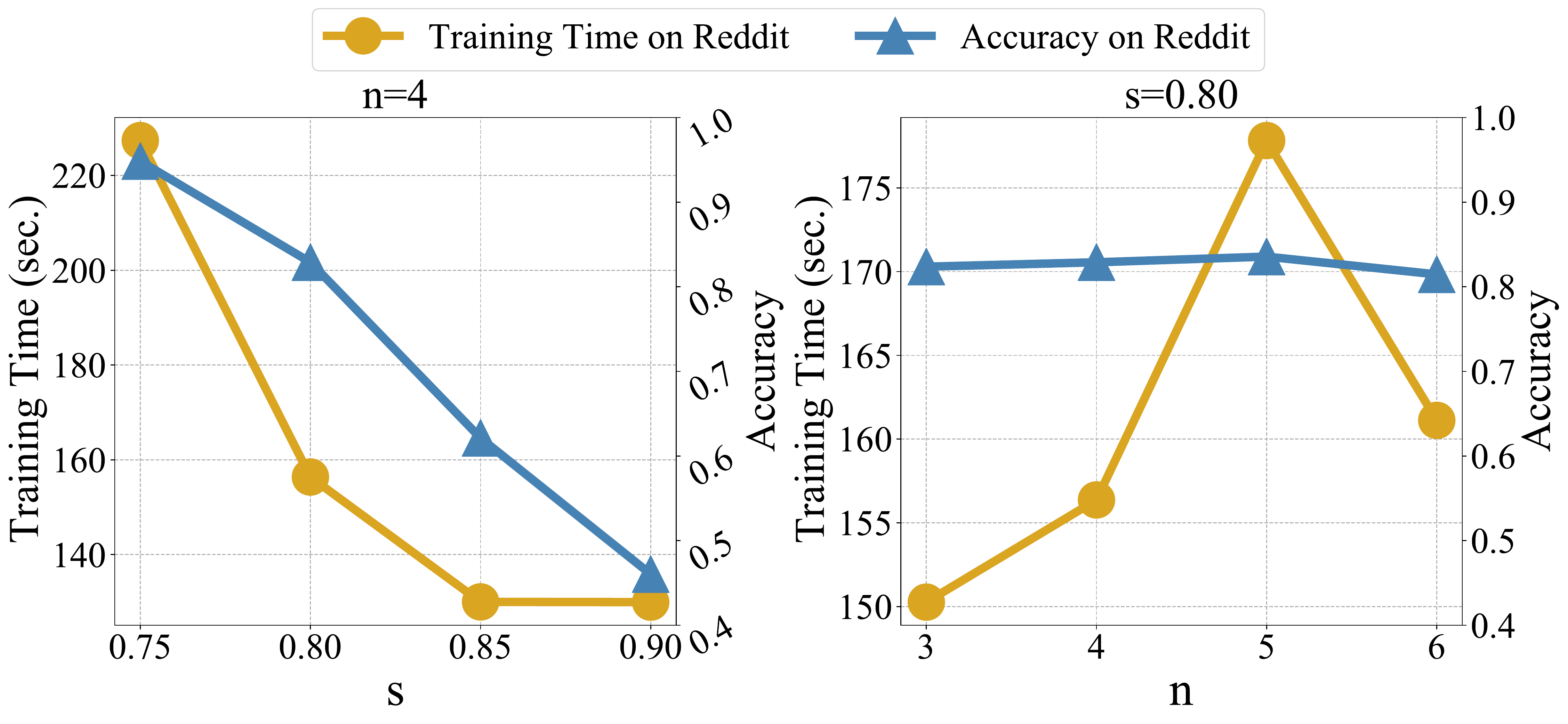}
%\caption{fig1}
\end{minipage}%
}%
\subfigure[]{
\begin{minipage}[t]{0.25\linewidth}
\centering
\includegraphics[width=1\columnwidth]{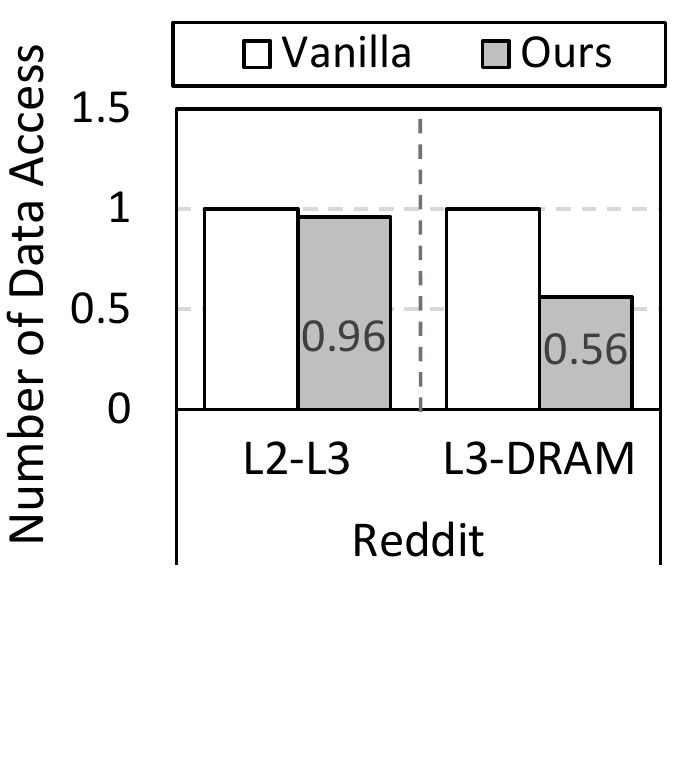}
%\caption{fig2}
\end{minipage}%
}%
\centering
\caption{(a) Exploration of the trade-off between training time and accuracy in GraphSAINT on Reddit by adjusting \texttt{n} \& \texttt{s}. (b) Comparison of the number of data access.}
\label{fig10}
\end{figure}

\noindent $\bullet$ \textbf{A good trade-off between training time and accuracy can be achieved by adjusting parameters \texttt{n} and \texttt{s}}. Since we merely retain the ``good nodes'' for sampling, our target is to find a trade-off point with considerable time reduction and tolerable accuracy loss.
As illustrated in the left subplot of Figure \ref{fig10}(a), as \texttt{s} increases (with \texttt{n} fixed), we have a tighter restriction on the quality of locality among neighbors per node, causing the nodes to be sampled are a minor part of total. This leads to accuracy loss since reducing nodes implies losing connections in a graph. Moreover, we sample nodes with good locality to reduce irregular memory access, indirectly saving training time.
In the right subplot of Figure \ref{fig10}(a), as $\texttt{n}$ increases (with \texttt{s} fixed), the training time is generally increasing before reaching a peak. When \texttt{n} is set to 5, we can obtain a competitive accuracy with an undesirable training time, implying a compromise of choosing a smaller \texttt{n} is acceptable under variations in the accuracy are trivial. Thus, there is a correlation between training time and accuracy. By adjusting the parameters, one can derive a comparable accuracy with an acceptable training time.

\begin{figure}[t]
\centering
\includegraphics[width=0.9\columnwidth]{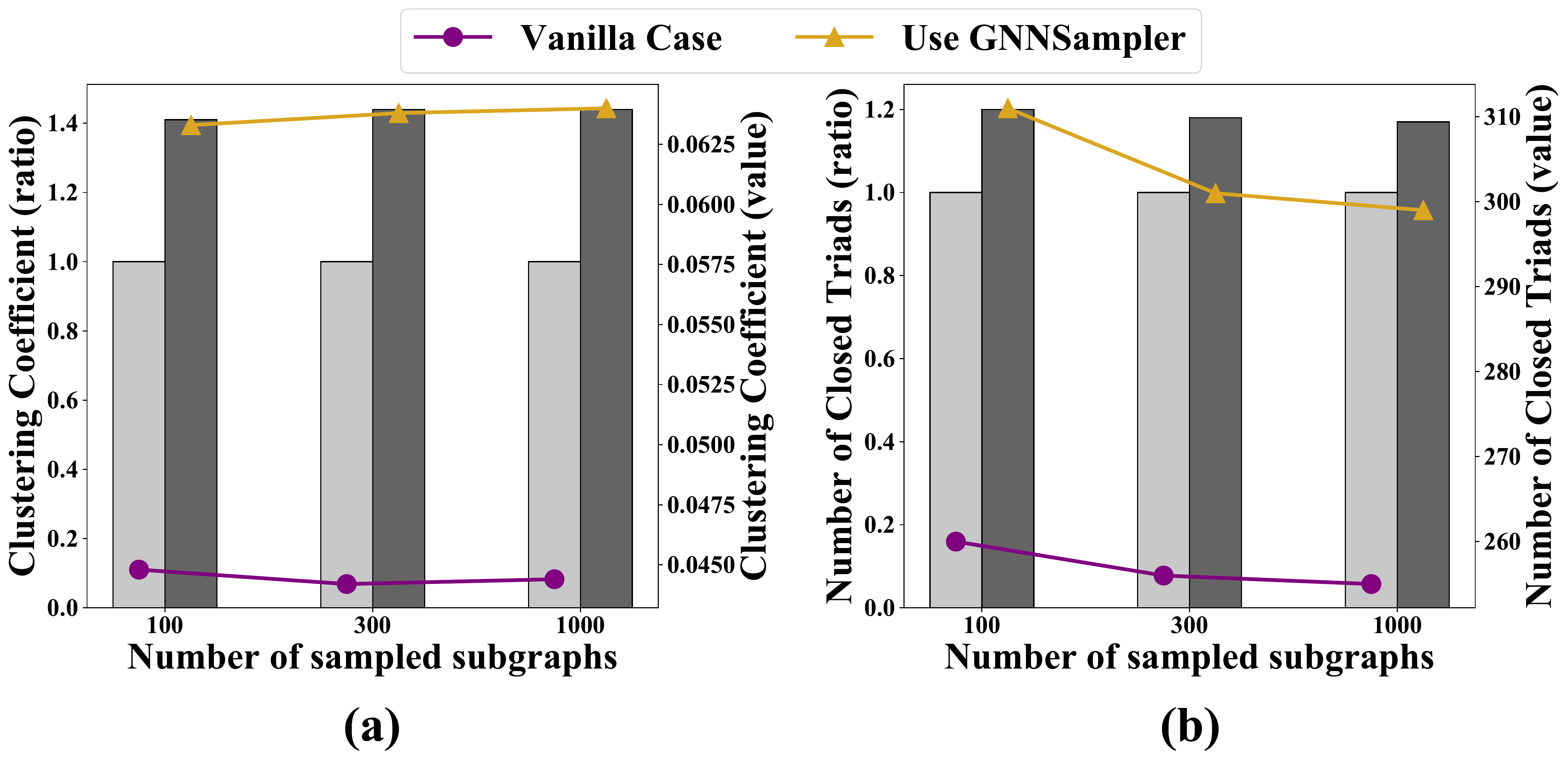}
\caption{Comparisons between vanilla and our (use GNNSampler) cases on Reddit among various metrics: (a) Comparisons on clustering coefficient; (b) Comparisons on the number of closed triads. Curves in subplots denote value variations of metrics.}
\label{fig:SNAP}
\end{figure}

\noindent $\bullet$ \textbf{Alleviating irregular memory accesses to neighbors helps reduce the training time.} As shown in Figure \ref{fig10} (b), we quantify the number of data access from L2 cache to L3 cache (L2-L3) and L3 cache to DRAM (L3-DRAM) with Intel PCM Tools \cite{PCM} to analyze the sampling process. Definitely, locality-aware optimizations can significantly reduce the number of data access in L3-DRAM under the condition that L2-L3 is almost similar. By reducing data access to DRAM, time of sampling is saved, which eventually accelerates the training.

\noindent $\bullet$ \textbf{Locality can be empirically reflected by the topology of sampled subgraphs.} With locality-aware optimizations, we argue that subgraphs sampled are more concentrated in the local structure, avoiding irregular or highly stochastic pattern in the graph topology. To reflect such properties, we conduct analysis via Stanford Network Analysis Platform (SNAP) \cite{Snap} tools. Specifically, by quantifying two metrics, i.e., the clustering coefficient (CC) and the number of closed triads (NCT), we analyze sampled subgraphs on Reddit using GraphSAINT model. 
CC is a local property that measures the cliquishness of a typical neighbourhood in a graph \cite{small-world}. In Figure \ref{fig:SNAP} (a), as the number of sampled subgraphs increases, the average CC of Ours is 1.42X of Vanilla, implying our sampled subgraphs are more concentrated and have a higher tendency of clustering.
NCT is a typical structure among three nodes with any two of them connected, which is used to reflect to a balanced group pattern in social networks \cite{closed_triads}. In Figure \ref{fig:SNAP} (b), NCT of Ours is 1.18X of Vanilla, indicating highly interconnected structures are sampled under locality-aware optimizations.

\section{Conclusion}
In this paper, we propose a unified programming model for mainstream GNN sampling algorithms, termed GNNSampler, to bridge the gap between sampling algorithms and hardware. Then, to leverage hardware features, we choose locality as a case study and implement locality-aware optimizations in algorithm level to alleviate irregular memory access in sampling. Our work target to open up a new view for optimizing sampling in the future works, where hardware should be well considered for algorithm improvement.

\section*{Acknowledgment}
This work was partly supported by the Strategic Priority Research Program of Chinese Academy of Sciences (Grant No. XDA18000000), National Natural Science Foundation of China (Grant No.61732018 and 61872335), Austrian-Chinese Cooperative R\&D Project (FFG and CAS) (Grant No. 171111KYSB20200002), CAS Project for Young Scientists in Basic Research (Grant No. YSBR-029), and CAS Project for Youth Innovation Promotion Association.

%\bibliographystyle{splncs04}
%\bibliography{mybibliography}

\end{document}